# Towards Human Cognition Level-based Experiment Design for Counterfactual Explanations (XAI)

Muhammad Suffian
*Dept. of Pure and Applied Sciences*
*University of Urbino 'Carlo Bo'*
Urbino, Italy
m.suffian@campus.uniurb.it

Muhammad Yaseen Khan
*Data Science*
*Daraz–Alibaba Group*
Karachi, Pakistan
muhammadyaseen.kha@alibaba-inc.com

Alessandro Bogliolo
*Dept. of Pure and Applied Sciences*
*University of Urbino 'Carlo Bo'*
Urbino, Italy
alessandro.bogliolo@uniurb.it

*Abstract*—Explainable Artificial Intelligence (XAI) has recently gained a swell of interest, as many Artificial Intelligence (AI) practitioners and developers are compelled to rationalize how such AI-based systems work. Decades back, most XAI systems were developed as knowledge-based or expert systems. These systems assumed reasoning for the technical description of an explanation, with little regard for the user's cognitive capabilities. The emphasis of XAI research appears to have turned to a more pragmatic explanation approach for better understanding. An extensive area where cognitive science research may substantially influence XAI advancements is evaluating user knowledge and feedback, which are essential for XAI system evaluation. To this end, we propose a framework to experiment with generating and evaluating the explanations on the grounds of different cognitive levels of understanding. In this regard, we adopt Bloom's taxonomy, a widely accepted model for assessing the users' cognitive capability. We utilize the counterfactual explanations as an explanation-providing medium encompassed with user feedback to validate the levels of understanding about the explanation at each cognitive level and improvise the explanation generation methods accordingly.

## I. INTRODUCTION

With the advent of Machine Learning (ML) as one of the forms of Artificial Intelligence (AI), we have seen remarkable outcomes which have replaced humans in critical decision-making. However, of the abstruse nature of these ML-based predictive models, in some sense, analogized to a black box, a motion of disquietude is prevailing that demands and necessitates the explanations of such decision-making for the basic comprehension and reliability of the predictive models. Developments in one domain can create an impact that may overlap with the observations made in other fields; we consider this phenomenon more conspicuous in cognitive science and AI. A plethora of explanation techniques have been devised, and their explanations are maintained under the necessity of users' understanding and trust in AI systems [1]. Developing explanations is a complex process; it combines technical (ML capabilities; model, features, hyper-parameters selection), human (users' and businesses' requirements, objectives, and information-processing capacities), and socio-technical (regulations, norms, processes, and stakeholders) perspectives.

Although a variety of methods in the field of Explainable Artificial Intelligence (XAI systems) are available, none of them is a panacea that can satisfy all diverse expectations and competing objectives of the users.

Under the notion of a social factor [2], many XAI systems exist that lag in providing explanations with merit, which can agitate the end-users and may result in gaining no trust in AI and XAI systems. Many researchers such as Doshi- Velez and Kim [3], Mueller *et al.* [4], Hoffman *et al.* [5], Miller [6], and Ribera and Lapedriza [7] have indicated that explanation is not a quality of words but rather a situational interaction between users and the AI system. Utilization of such interactions (i.e., feedback from the end-users) would also help in the designing and refining of the XAI systems; at the same time, such an interactive environment helps the end-users to understand the explanation per their cognitive capabilities and motivates them to embody thoughts into actions [8]. A comprehensive set of requirements that pertain to XAI's technical characteristics, such as providing human-readable output, has been put out by several researchers [4], [7], [9]–[11]. However, there is currently a significant gap in gathering and utilizing end-user insights and feedback in designing XAI systems, as a few studies have addressed this subject [12]– [14]. In addition, the XAI research focus seems to have shifted to a more realistic or pragmatic account of understanding [15]. An extended area where cognitive science research may significantly impact XAI developments for measuring user knowledge, feedback, and understanding are crucial for evaluating XAI systems. These measures might be employed to evaluate the efficacy of the explanations and provide the XAI system feedback to improve the explanations [16]. Hence, we maintain that the development of human-centred XAI systems: alongside the matters which qualify as an explanation, relies on the intersection of the users' needs, background knowledge, and goals/objectives; the requirement of human-in-the-loop is essential. These requirements entail cognitive-behavioural aspects and the structure of explanations.

Under interactive approaches for XAI, users are provided with explanations which take various organizational, cultural, and human context-dependent aspects into account [9], [17]. Although these requirements are necessary preconditions, they do not ensure that the explanations are aligned with the users' aims, demands, and operational environments. In this context, *counterfactual explanations*[1] (CE) are significant and have a tendency to deal with the complexity of the aforementioned problems. CE can involve users in the explanation generation mechanism and thus able to generate explanations complying with user-defined constraints.

For example, in the attempt to decide why one should receive a loan? The recipient should be provided with a justification for the predicted outcome [18]. However, with the counterfactual explanation, it is worth considering the actions that can dwindle the possibility of experiencing the same results by the recipient. Thus for the interaction, the very process of the explanation would suggest the end-user choosing the optimum actions, which include proposing adequate and admissible value for features for addressing the question.

One of the well-qualified pre-existing frameworks of human cognition and assessments thereof is Bloom's taxonomy [19]; we adopt the revised Bloom's taxonomy [20]. In this study, we consider human[2] as an end-user for whom the explanation is concerned.

This paper presents a framework for generating and evaluating explanations based on different cognitive levels of understanding. In this regard, we categorize the stakeholders into two categories: domain experts and novice users. We employ three cognition levels:

- Knowledge (it would help to recall the information about the facts and concepts).
- Comprehension (meaning from the facts and examples).
- Application (ability to apply the learned information for generating new explanations).

We utilize CE as an explanation-providing medium encompassed with user feedback to validate the levels of understanding about explanation at each cognitive level. User feedback will be utilized to customize the explanations and find commonalities that could simplify the general explanations for all cognitive levels and specific explanations for specific cognitive levels. We also record human behaviours as different levels of cognitive behaviours. The proposed framework serves as the guidelines for designing and developing XAI systems, and its domain-specific implementation would help to initiate a dialogue or interaction with the systems, increasing the interest and trust of the users. The successful evaluation of this experiment will initiate a new line of research for the XAI community to encompass and categorize explanations as per cognitive levels of humans; and understandable explanations from a broader perspective.

## II. BACKGROUND

**Perspective of explanation in different scientific domains.** Classically, the explanations are supposed to be divided into three types/considerations: (i) as a 'deductive proof' such as in work maintained by Hempel and Oppenheim [21], (ii) in terms of 'causality' of some event as maintained in the work of Verma and Pearl [22], [23], and (iii) as a 'mental model', for which Allen [24] maintained a neuro-scientific reference that it is internal representations, which emerge when a particular region of the brain is active (see [25] as well).

Hoffman *et al.* [26] offered a conceptual model of an XAI system's explanation processes that are based on human explanation processes, explaining how a user's logical understanding of the XAI system's explanations may be assessed at different functional phases. Hoffman *et al.* [Ibid.] claims that the effort necessary to develop a mental model may be utilized to determine how well a user comprehends the XAI system.

In the context of social sciences, Miller [27] maintained that human explanation processes might be divided into four categories: (i) explanation definition, (ii) generation, (iii) selection, and (iv) evaluation. Feltovich *et al.* [28] and Mueller and Tan [29] highlighted how 'beliefs'—including false beliefs cause hindrance in seeking a change or accepting explanations. Another study has focused more directly on explanation, particularly about the 'illusion of explanatory depth', in which individuals believe that they understand something to a sufficient degree and thus exaggerate their (mis)understanding [29]–[31].

**Interaction and Cooperation in the process of explanation.** An explanation could be seen as the comprehension of instructional content given to the learner; however, for AI-based automated decision systems, it is interaction or co-adaptation between the user and the system. Bre'zillon and Pomerol [32] foregrounded that cooperation is enhanced by explanation and, in a similar relationship, a pertinent explanation would be possible to achieve through cooperation.

In addition, the research on seeking better natural explanations has led to several discoveries; from there, we also maintain that conversation is a fundamental cooperative activity and significantly helpful for explanations [33]. In the same context, scientific evidence from studies on natural explanation generally acknowledges that explanations and human-to-human communication, in general, are co-adaptive processes that necessitate both the explanation and the learner's ability to see things from the perspective of the other. It means that both the explanation and the learner have internalised mental representations.

**User preferences in explanatory reasoning.** On the explanations generated by the XAI system, some studies have reported variations among the end-users w.r.t the matter of satisfaction and preferences of selection of the type of explana- tion. In the same context, we see the reason for such variations is distinctive in terms of personality traits or thinking patterns. Of the satisfaction, people tend to prefer:

- A brief or cursory justification that includes fewer causal references [34]. However, a subsequent failure of such preference is the 'illusion of explanatory depth', which is the self-assertive idea that one's comprehension is adequate [35].
- A thoughtful and contemplative answers because these are more satisfying [36].

---

[1]From here onwards, we shall use explanation(s) and counterfactual explanation(s) as interchangeable terms.

[2]In rest of the paper, we use human, user, end-user, information seeker, and learner interchangeably.

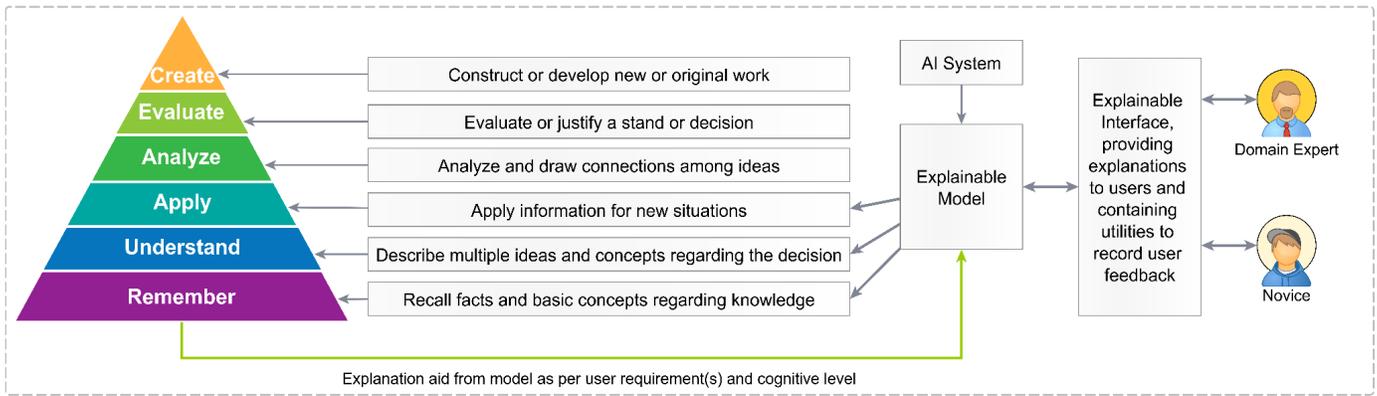

Fig. 1. Explainable Artificial Intelligence and Human-Centred Perspective.

- Complex explanations over simple ones. Complicated explanations differ significantly among cultures [37]—however, the desire is for simplicity.

## III. PROPOSED FRAMEWORK

This section describes the proposed experimental design based on human cognition levels to categorize and evaluate the explanations as human-centred. It deals with two types of users, *i.e.*, Novice and domain expert; the segregation thereof is itemized below:

- **Novice** is supposed to have limited knowledge about the domain, AI, and XAI system; a lower cognitive level per Bloom's taxonomy; however, the cognition of the novice *in esse* is lower than its counterpart. The actual meaning of an XAI system is to serve the explanation for the user in simplified forms; thus, if a novice thinks that the explanation is not comprehensible, she can initiate a request for generating the alternate/modified explanation in a more simplified manner.
- **Domain Expert** is supposed to have a higher cognition level per Bloom's taxonomy. Such users have sound knowledge about the domain, understanding of the matter, inputs, predictive outcomes, and the explanations generated for the instances. Forbye it, if the generated explanations are not up to the standards, the domain expert is the principal stakeholder for rejecting such explanations; likewise, the Novice, the domain expert, may ask for explanations in a more simplified form.

The architecture of the proposed framework is illustrated in figure 1, where we can see that Bloom's taxonomy is labelled with definitions of its different cognitive levels, and an AI system for which the explanations are required, an explainable model that generates counterfactual explanations, and explainable interface with the in-built utilities for providing explanations and recording the respective feedback to/from end-users. We focus on the three foundational levels of Bloom's taxonomy, which are *knowledge*, *comprehension*, and *application*—for referencing the cognitive level of the novice user and modifications made in the explanation generation mechanism. A green arrow from the pyramid of Bloom's taxonomy is thus directed towards the explainable model.

Incrementally, the experiment will generate the set of explanations for a single instance—keeping the domain/context and user-defined constraints into account—and presenting to both end-users. For the given explanation, the response of the end-users would be dichotomous, *i.e.*:

1) *Agreement*, which would mean that the given explanation was satisfactory and comprehensible by the respective stakeholder. The system will record a positive evaluation on the respective instance.
2) *Disagreement* would mean that the generated explanation is not up to the mark; consequently, a negative evaluation is recorded. In this scenario:
   - If the end-user is a novice, then (under the impression of improper understanding) the system will follow an exploration of cognitive learning mechanism utility[3], and alongside it, an 'application' utility[4].
   - Otherwise, for the domain expert, it would be more interesting to follow the 'application' utility.

After taking a tour of the cognitive learning mechanism and application utilities, users can return to the main process and mark the previously presented explanation with their agreement or disagreement.

Meanwhile, in the interactive process, some cases could arise that are based on the agreement of the presented explanations. These cases are illustrated in figure 2 and discussed in the subsequent paragraphs.

---

[3]The system will initiate an interactive utility that is based on Bloom's cognitive learning mechanism, where the user can revise/gain knowledge about the facts and concepts used in the previously presented explanation, and understand the key elements/indicators of the explanation.

[4] It has the opportunity to make changes in the input of the system (as the counterfactual XAI system offers such functionality to observe the dynamics of the reasoning and underlying domain-specific factors) and observe the behaviour of the outcomes.

TABLE I
QUESTIONNAIRE FOR THE END-USERS TO EXPLORE COGNITIVE LEVEL INFORMATION.

| Cognition Level | Questions | Counterfactual XAI systems as Decision-Aid |
| --- | --- | --- |
| Knowledge | <ul><li>What is the decision-making process?</li><li>What are the different terms, and what do they mean?</li><li>What are the important concepts to know?</li><li>What are the basic features of/in the data?</li><li>How is the decision made?</li></ul> | XAI system would offer end-users individualized training (knowledge) about the domain terms, concepts, and facts utilized in the explanations. It will provide the answers to user-selected questions through interaction utility. |
| Comprehension | <ul><li>Why does the system provide this output?</li><li>What drives its decision, and why?</li><li>What governing rules and domain-specific variables had a role in the decision?</li><li>Are specific indications that make an explanation necessary rather than whether it should be provided consistently [38]?</li></ul> | XAI system will deliver examples and actionable changes to understand the behaviour of the predictive model, the hard constraints, and the implausible boundaries. Ultimately enabling end-users to learn how the desired outcome could be achieved. The XAI system may concentrate on particular aspects of the issue or offer a comprehensive summary, depending on the end-user's needs. |
| Application | <ul><li>How/Why was this decision made rather than others?</li><li>How/Why would the system respond if the input is altered with X features?</li><li>What happens if the X features/parameters are protected?</li></ul> | XAI system will allow the end-users to make changes in the model's input and observe outcomes coherent with examples and actionable suggestions provided at the previous levels. |

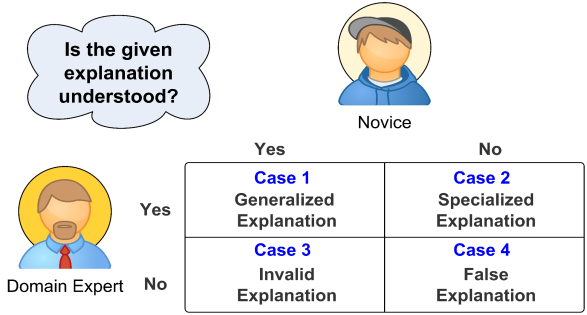

Fig. 2. Categorization of explanations based on user feedback.

**Case 1: Generalized Explanations,** *i.e.*, the explanations are satisfiable and comprehensible by both end-users. This unanimous acceptance stamps that the explanation was enriched with enough knowledge and facts that even novice users could understand it, and the agreement from the domain expert reflects the applicability of the explanation in terms of higher fidelity. Consequently, the system will record a positive evaluation score for the underlying test instance.

**Case 2: Specialized Explanations,** *i.e.*, the only domain expert agreed/satisfied with explanations. It infers that an explanation consists of more sophisticated and domain-specific knowledge that novice user is unable to understand. Therefore, we can not recognize this explanation as generally accepted by all users but rather as a specialized explanation for the domain expert. It would be the event for moving towards cognitive learning and application utilities for the novice. The utility of the cognitive learning mechanism provides potential queries (see table I) regarding all aspects of the explanation, such as the required knowledge to understand the behaviour of the underlying explanation mechanism.

**Case 3: Invalid Explanation,** *i.e.*, only novice agreed/satisfied with explanations. In a sceptical motion, this case raises some questions about the credibility of the explanation. Since the domain expert disagrees with the presented reasoning, it draws some confusion about whether the underlying ML model is accurate or not. For example, the underlying predictive model provides desired results (favourable to the novice) on a specific test instance with certain input parameters. Although the explanation method tries to explain this wrong outcome, according to a domain expert, it is an error and a biased or misclassified output. Thus, vetoed by the domain expert, a revised explanation will be provided.

**Case 4: False Explanations,** *i.e.*, no one is satisfied with explanations. Technically, the explanation is not coherent with actual input and probably trying to explain the wrong facts. Technically, the explanation is not coherent with actual input and probably trying to explain the wrong facts. Some other aspects could be behind these explanations, such as the biased underlying prediction model and implausible counterfactual instance. The former could force the explainer to explain wrong predictions. The latter suggests extreme modifications which are not acceptable to a novice and, technically, violate data distribution to whom domain-expert refuse to accept. A negative evaluation will be stored for such an explanation. In this case, the explanation generation methods need to be examined and reformed.

During the interaction with the utility of learning, a questionnaire will assist the end-users in comprehending and evaluating the domain/information at each cognitive level. The potential questions of end-users are categorized according to cognitive levels and their expected answers from the explaining model in table I. Case 1 and case 2 evaluate the 'goodness' of our framework. The explanations are better if they belong to these two cases.

## IV. DISCUSSION AND CONCLUSION

Studies, which are made on the overlapping fields of cognitive science and AI, have strikingly impacted XAI system designs. The pivot of such studies is considering human-in-the-loop and, thus, examining the structures and methods of improvising the effectiveness explanations from the perspective of humans. The purpose of explanation is to reduce the knowledge gap between the explainer and the explainee. In this regard, the explanation should consider the context and user understanding, such as which sort of end-user is looking for an explanation. And for what matters, and with which means is the explanation supposed to deal with questions asked by end-users? Forbye it, explanations must be goal-oriented, for which we maintain that it should assist the end-user in taking the proper steps in comprehending the AI system.

XAI systems are evaluated based on how well their explanations achieve the goal of the human-in-the-loop. Bloom's taxonomy provides another lens through which we can evaluate stakeholders' cognitive capabilities for understanding the AI and XAI systems. We propose an experimental design to evaluate the explanations according to various cognitive levels of learning. To validate the amount of understanding of an explanation at each cognitive level, we employ counterfactual explanations as an explanation-providing medium that is further evaluated through user feedback. Utilizing user feedback will help to tailor the explanations as specialized and generalized based on cognitive levels. The appraisal of the proposed experiment will endorse a new line of research with human cognition-level design in enabling end-users to comprehend explanations better and making the explanations more satisfactory and trustworthy.

## Acknowledgment

This work is supported by REMEST at the Department of Pure and Applied Sciences, University of Urbino Carlo Bo, Urbino, Italy.

## References

[1] B. Paranjape, J. Michael, M. Ghazvininejad, L. Zettlemoyer, and H. Hajishirzi, "Prompting contrastive explanations for commonsense reasoning tasks," *arXiv preprint arXiv:2106.06823*, 2021.

[2] A. Holzinger, A. Saranti, C. Molnar, P. Biecek, and W. Samek, "Explainable ai methods-a brief overview," in *International Workshop on Extending Explainable AI Beyond Deep Models and Classifiers*, pp. 13–38, Springer, 2022.

[3] F. Doshi-Velez and B. Kim, "Towards a rigorous science of interpretable machine learning," *arXiv preprint arXiv:1702.08608*, 2017.

[4] S. T. Mueller, E. S. Veinott, R. R. Hoffman, G. Klein, L. Alam, T. Mamun, and W. J. Clancey, "Principles of explanation in human-ai systems," *arXiv preprint arXiv:2102.04972*, 2021.

[5] R. R. Hoffman, S. T. Mueller, G. Klein, and J. Litman, "Met- rics for explainable ai: Challenges and prospects," *arXiv preprint arXiv:1812.04608*, 2018.

[6] T. Miller, "Explanation in artificial intelligence: Insights from the social sciences," *Artificial intelligence*, vol. 267, pp. 1–38, 2019.

[7] M. Ribera and A. Lapedriza, "Can we do better explanations? a proposal of user-centered explainable ai.," in *IUI Workshops*, vol. 2327, p. 38, 2019.

[8] M. Miłkowski and M. Hohol, "Explanations in cognitive science: unification versus pluralism," *Synthese*, vol. 199, no. 1, pp. 1–17, 2021.

[9] F. Cabitza and J.-D. Zeitoun, "The proof of the pudding: in praise of a culture of real-world validation for medical artificial intelligence," *Annals of translational medicine*, vol. 7, no. 8, 2019.

[10] S. T. Mueller, E. S. Veinott, R. R. Hoffman, G. Klein, L. Alam, T. Mamun, and W. J. Clancey, "Principles of explanation in human-ai systems," *arXiv preprint arXiv:2102.04972*, 2021.

[11] R. K. Mothilal, A. Sharma, and C. Tan, "Explaining machine learning classifiers through diverse counterfactual explanations," in *Proceedings of the 2020 conference on fairness, accountability, and transparency*, pp. 607–617, 2020.

[12] M. Suffian, P. Graziani, J. M. Alonso, and A. Bogliolo, "FCE: Feedback based counterfactual explanations for explainable ai," *IEEE Access*, vol. 10, pp. 72363–72372, 2022.

[13] M. Naiseh, N. Jiang, J. Ma, and R. Ali, "Personalising explainable recommendations: literature and conceptualisation," in *World Conference on Information Systems and Technologies*, pp. 518–533, Springer, 2020.

[14] L. Sanneman and J. A. Shah, "A situation awareness-based framework for design and evaluation of explainable ai," in *International Workshop on Explainable, Transparent Autonomous Agents and Multi-Agent Systems*, pp. 94–110, Springer, 2020.

[15] A. Pa´ez, "The pragmatic turn in explainable artificial intelligence (xai)," *Minds and Machines*, vol. 29, no. 3, pp. 441–459, 2019.

[16] R. R. Hoffman, S. T. Mueller, G. Klein, and J. Litman, "Met- rics for explainable ai: Challenges and prospects," *arXiv preprint arXiv:1812.04608*, 2018.

[17] C. M. Cutillo, K. R. Sharma, L. Foschini, S. Kundu, M. Mackintosh, and K. D. Mandl, "Machine intelligence in healthcare—perspectives on trustworthiness, explainability, usability, and transparency," *NPJ digital medicine*, vol. 3, no. 1, pp. 1–5, 2020.

[18] W. F. Taylor, "Meeting the equal credit opportunity act's specificity requirement: Judgmental and statistical scoring systems," *Buff. L. Rev.*, vol. 29, p. 73, 1980.

[19] D. R. Krathwohl, "A revision of bloom's taxonomy: An overview," *Theory into practice*, vol. 41, no. 4, pp. 212–218, 2002.

[20] L. O. Wilson, "Anderson and krathwohl–bloom's taxonomy revised," *Understanding the New Version of Bloom's Taxonomy*, 2016.

[21] C. G. Hempel and P. Oppenheim, "Studies in the logic of explanation," *Philosophy of Science*, vol. 15, no. 2, p. 135–175, 1948.

[22] T. S. Verma and J. Pearl, "Equivalence and synthesis of causal models," in *Probabilistic and Causal Inference: The Works of Judea Pearl*, pp. 221–236, 2022.

[23] J. Pearl, *Causality*. Cambridge university press, 2009.

[24] R. B. Allen, "Mental models and user models," in *Handbook of human-computer interaction*, pp. 49–63, Elsevier, 1997.

[25] T. Kulesza, S. Stumpf, M. Burnett, S. Yang, I. Kwan, and W.-K. Wong, "Too much, too little, or just right? ways explanations impact end users' mental models," in *2013 IEEE Symposium on visual languages and human centric computing*, pp. 3–10, IEEE, 2013.

[26] R. R. Hoffman, G. Klein, and S. T. Mueller, "Explaining explanation for "explainable ai"," in *Proceedings of the Human Factors and Ergonomics Society Annual Meeting*, vol. 62, pp. 197–201, SAGE Publications Sage CA: Los Angeles, CA, 2018.

[27] T. Miller, "Explanation in artificial intelligence: Insights from the social sciences," *Artificial intelligence*, vol. 267, pp. 1–38, 2019.

[28] P. J. Feltovich, R. L. Coulson, and R. J. Spiro, "Learners'(mis) understanding of important and difficult concepts: A challenge to smart machines in education," *Smart machines in education*, pp. 349–375, 2001.

[29] S. T. Mueller and Y.-Y. S. Tan, "Cognitive perspectives on opinion dynamics: The role of knowledge in consensus formation, opinion divergence, and group polarization," *Journal of Computational Social Science*, vol. 1, no. 1, pp. 15–48, 2018.

[30] L. Rozenblit and F. Keil, "The misunderstood limits of folk science: An illusion of explanatory depth," *Cognitive science*, vol. 26, no. 5, pp. 521–562, 2002.

[31] N. Rabb, J. J. Han, and S. A. Sloman, "How others drive our sense of understanding of policies," *Behavioural Public Policy*, vol. 5, no. 4, pp. 454–479, 2021.

[32] P. Bre´zillon and J.-C. Pomerol, "Contextual knowledge sharing and cooperation in intelligent assistant systems," *Le travail humain*, pp. 223– 246, 1999.

[33] R. W. Southwick, "Explaining reasoning: an overview of explanation in knowledge-based systems," *The knowledge engineering review*, vol. 6, no. 1, pp. 1–19, 1991.

[34] T. Lombrozo, "Simplicity and probability in causal explanation," *Cognitive psychology*, vol. 55, no. 3, pp. 232–257, 2007.

[35] L. Rozenblit and F. Keil, "The misunderstood limits of folk science: An illusion of explanatory depth," *Cognitive science*, vol. 26, no. 5, pp. 521–562, 2002.

[36] S. A. Sloman, P. M. Fernbach, and S. Ewing, "A causal model of intentionality judgment," *Mind & Language*, vol. 27, no. 2, pp. 154–180, 2012.

[37] G. Klein, L. Rasmussen, M.-H. Lin, R. R. Hoffman, and J. Case, "Influencing preferences for different types of causal explanation of complex events," *Human factors*, vol. 56, no. 8, pp. 1380–1400, 2014.

[38] W. Kintsch, "The representation of knowledge and the use of knowledge in discourse comprehension," in *North-Holland Linguistic Series: Linguistic Variations*, vol. 54, pp. 185–209, Elsevier, 1989.